\documentclass[nameyear]{article}

\usepackage[preprint]{nips_2018}
\usepackage{framed,multirow}
\usepackage{algorithm}
\usepackage{algpseudocode}

\usepackage{amssymb}
\usepackage{amsmath}
\usepackage{latexsym}
\usepackage{graphicx}
\usepackage{subcaption}

\usepackage{url}
\usepackage{xcolor}
\definecolor{newcolor}{rgb}{.8,.349,.1}
\def\x{\mathbf{x}}
\def\y{\mathbf{y}}

\renewcommand{\gamma}{\mathbf{T}}

\let\emph\textit

\title{An Entropic Optimal Transport Loss for Learning Deep Neural Networks under Label Noise in Remote Sensing Images}

\author{Bharath Bhushan Damodaran \thanks{Under consideration at Computer Vision and Image Understanding.} \\
  Universit\'e de Bretagne Sud\\ IRISA, UMR 6074, CNRS\\          Vannes-56000, France\\
  \texttt{bharath-bhushan.damodaran@irisa.fr} \\
   \And
  R\'emi Flamary \\
  Universit\'e C\^ote d'Azur \\
   Lagrange, UMR 7293, CNRS, OCA \\
    Nice-06108, France \\
   \texttt{remi.flamary@unice.fr} \\
   \AND
  Viven Seguy \\
  Kyoto University \\
   Japan\\
   \texttt{vivien.seguy@iip.ist.i.kyoto-u.ac.jp} \\
   \And
  Nicolas Courty \\
  Universit\'e de Bretagne Sud\\ IRISA, UMR 6074, CNRS\\          Vannes-56000, France\\
   \texttt{nicolas.courty@irisa.fr} \\
  }
\begin{document}
\maketitle

\begin{abstract}
Deep neural networks have established as a powerful tool for large scale supervised classification tasks. The state-of-the-art performances of deep neural networks are conditioned to the availability of large number of accurately labeled samples. In practice, collecting  large scale accurately labeled datasets is a challenging and tedious task in most scenarios of remote sensing image analysis, thus cheap surrogate procedures are employed to label the dataset. Training deep neural networks on such datasets with inaccurate labels easily overfits to the noisy training labels and degrades the performance of the classification tasks drastically. 
To mitigate this effect, we propose an original solution with entropic optimal transportation. It allows to learn in an end-to-end fashion deep neural networks that are, to some extent, robust to inaccurately labeled samples. We empirically demonstrate on several remote sensing datasets, where both scene and pixel-based hyperspectral images  are considered for classification. Our method proves to be highly tolerant to significant amounts of label noise and achieves favorable results against state-of-the-art methods.

\end{abstract}


\section{Introduction}
\label{sec1}
Deep learning has been applied with tremendous success on a variety of tasks in remote sensing image analysis. For instance, achievement of state-of-the-art performance in scene classification \citep{Cheng2018,Anwer2018}, pixel-wise labeling of both multispectral \citep{Huang2018,Audebert2018,Maggiori2017} and hyperspectral datasets \citep{Zhong2018,Wang2017}, object detection \citep{Kellenberger2018} and image retrieval \citep{Zhou2018,Ye2017,Li2018}, highlights the recent success of deep learning models in remote sensing. But these phenomenal performances is highly dependant on the availability of large collection of datasets with accurate annotations (labels). If either the size of the dataset or the accuracy of the labels is not sufficient (i.e, small scale datasets or inaccurate labels), the performance of the deep learning methods could suffer drastically. 
The former one can be addressed to some degree by data augmentation strategies, however solving the later case of inaccurate labeling is more difficult.

To address the large scale data requirements of deep learning methods, new datasets have been proposed recently in the remote sensing community \citep{Zhou2018,Huang2018,Cheng2017,Kemker2017,Wang2016,Xia2017}. This trend will grow continuously in the coming years, due for instance to the large constellation of the Earth observation satellites. 
One of the major challenge in collecting this new large scale data is accurate labeling of the samples. Manual expert labeling of such large collection of samples is often not feasible and not cost-effective. Thus, labeling is usually performed by non-experts through crowd sourcing \citep{Snow2008,Haklay2010}, keyword query through search engine in the case of images, open street maps, and out-dated classification maps \citep{Kaiser2017}. These cheap surrogate procedures allows scaling the size of labeled datasets, but at the cost of introducing label noise (i.e. inaccurately labeled samples).
Even when manual experts are involved in labeling the data samples, they must be provided with sufficient information; otherwise inaccurate labeling may still occur (for instance, during the field survey) \citep{Hickey1996}. Note that in the some applications, labeling is a subjective task \citep{Smyth1995} that can again introduce label noise. Furthermore, the label noise could occur due to the misregistration of satellite images. Hence in general, large scale datasets might mostly contain inaccurately labeled samples or affected by label noise. In this case, when deep learning methods are employed with conventional loss functions (for instance, categorical cross entropy, mean square error), they will not be robust to label noise, and as a result the classification accuracy decreases significantly \citep{Zhang2017}. This calls for robust approaches to mitigate the impact of label noise on the deep learning methods. 


Recently, it was shown that while training deeper neural networks, models tend to memorize the training data, and this phenomena is more severe when the dataset is affected by the label noise \citep{Zhang2017}. The impact of the label noise in the deep learning models can be partly circumvented by regularization techniques such as drop out layers, and weight regularization. These standard procedures make neural networks robust to some extend, but they are still prone to memorize noisy labels for medium-to-large noise levels. The problem of learning with noisy labels has been long studied in machine learning \citep{Frenay2014,Brooks,Zhu2004a,Saez2014,Hickey1996,Smyth1995,Natarajan2013}, but still only few works have focused on neural networks. Recently, new approaches have been proposed in the computer vision and machine learning fields to tackle the label noise by cleaning the noisy labels or designing robust loss functions within the deep learning framework \citep{Jiang2018,Vahdat2017,Patrini2017}.

To mitigate the impact of label noise, one category of method relies on estimating the noise transition probability that describes the probability of $i^{th}$ class label being mislabeled to the $j^{th}$ class label, 
 and use it to be robust to label noise \citep{Vahdat2017,Natarajan2013,Patrini2017}. Among those, some of them require a small set of clean labels to estimate the noise transition probability \citep{Vahdat2017}. The other category of methods proposes to use loss functions which are inherently tolerant to the label noise \citep{Natarajan2013,Rooyen2015,Masnadi-ShiraziHamedandVasconcelos2008,Ghosh2015,Aritra2017}. Though these methods provided satisfactory results, none of them consider the implicit local geometric structure of the underlying data.  

The primary objective of this paper is to develop a robust approach to tackle the label noise for remote sensing image analysis. The sensitiveness of deep neural networks to label noise has not been well studied in remote sensing image analysis so far as per our knowledge. Hence the first contribution of this article lies in studying the robustness of deep neural networks to label noise, and also to analyse the efficiency of existing robust loss functions for remote sensing classification tasks. 
The second contribution of this paper is to propose a novel robust solution to tackle the label noise based on optimal transportation theory~\citep{Villani09}. 
Indeed we propose to learn a deep learning model which is robust to label noise by fitting the model to the label-features joint distribution of the dataset with respect to the entropy-regularized optimal transport distance. We coin this method as CLEOT for Classification Loss with Entropic Optimal Transport. One major advantage of our approach compared to existing methods is that our method inherently exploits the geometric structure of the underlying data. A stochastic approximation schemes is proposed to solve the learning problem, and allows the use of our approach within deep learning frameworks. Experiments are conducted on several remote sensing aerial and hyperspectral benchmark datasets, and the results demonstrate that our approach is more robust (tolerant) to high level label noise than current state-of-the-art methods.

The remaining of the paper is organized as follows. Section \ref{sec2} discusses related works, section \ref{sec3} defines the label noise and describes the problem formulation, and section \ref{sec4} introduces optimal transport. The proposed method is then presented in section \ref{sec5} while experimental datasets and results are explained in section \ref{sec6}. We finally draw some conclusions in section \ref{sec7}.
  
\section{Related works}
\label{sec2}


\subsection{Learning with noisy labels}
Label noise, and attribute (feature) noise are two types of noise commonly found in machine learning datasets. The label noise is considered as more harmful and difficult to tackle compared to the attribute noise, and can decrease the classification performance significantly \citep{Zhu2004a}. Learning with noisy labels with shallow learning methods have been widely investigated in the literature \citep{Frenay2014,Brooks,Zhu2004a,Saez2014,Hickey1996,Smyth1995,Natarajan2013}, but studies in the context of deep learning still remain scarse (but growing recently) \citep{Reed2015,Vahdat2017,Hendrycks2018,Patrini2017}. Among the several methods which have been proposed to robustly train deep neural networks on the datasets with noisy labels, one set of methods approaches the problem from the perspective of cleaning the noisy labels, and use the clean estimated labels for training deep neural networks, or they smoothly reduce the impact of noisy labels by putting smaller weights on noisy label samples, either through directed graphical models \citep{TongXiao2015}, conditional random field \citep{Vahdat2017}, knowledge graph distillation \citep{Li2017}, meta-learning \citep{Ren2018} or noise-transition matrix estimation \citep{Hendrycks2018}. 
But those methods require an additional small subset of data with clean labels, or require ground truth of pre-identified noisy labels in order to model the noise in the dataset.
A second kind of methods tries to detect clean instances out of the noisy instances, and use them to update the parameters of the trained neural network \citep{Jiang2018, Ding2018}. In this category, two deep networks or two stage framework are employed to remove noisy label instances.
The last kind of methods design a robust loss function and loss correction approach. The robust loss functions unhinged \citep{Rooyen2015}, savage \citep{Masnadi-ShiraziHamedandVasconcelos2008}, sigmoid and ramp \citep{Ghosh2015} are inherently robust to the label noise with associated theoretical bounds. Most of these method rely on an assumption of symmetric loss function. The loss correction approaches employ the correctness procedure to adjust the loss function to eliminate the influence of the noisy labels by forward and backward correction approach \citep{Patrini2017} using the estimated noise transition model from the noisy labeled data, adding linear layer on top of a softmax layer \citep{Sukhbaatar2014,Jacob2017}, using bootstrap approach \citep{Reed2015} that replaces the noisy labels with a soft or hard combination of noisy labels and their predicted labels. 

In remote sensing image analysis, the adverse effect of the label noise is not much studied in literature. 
The impact of label noise has been recently studied in \citep{Frank2017,Pelletier2017} with shallow classifiers. The feasibility of using online open street map (outdated or mislabeled ground truth) to obtain classification map with deep neural network was studied in \citep{Kaiser2017}, however they didn't consider directly addressing label noise as a specificity of the problem. Some other studies tackle the label noise in the context of shallow classifier (random forest, logistic regression) by selecting clean labeled instances via outlier detection \citep{Pelletier2017b}, or by using existing noise tolerant logistic regression method \citep{Maas2016,Maas2017}. 

\subsection{Optimal transport}

Optimal transport theory provide the Wasserstein distance, that measures the discrepancy between probability distribution in a geometrically sound manner. More recently, optimal transport has found applications in  domain adaptation \citep{Cou17, courty2017joint, Damodaran2018}, generative models \citep{seguy2018, genevay2017sinkhorn, arjovsky17}, data mining \citep{Courty2017} and image processing \citep{Solomon2015,Papadakis2015}.

Among those applications, domain adaptation is the one that is the most related to the problem of noisy labels. It indeed aim at adapting a classifier to better predict on new data whose distribution is different from the training data. In the case of label noise in the training dataset, we want to adapt the classifier to perform well on data using a different noisy dataset for training. One recent approach coined JDOT for joint distribution optimal transport \citep{courty2017joint} propose to estimate a classifier that minimize the Wasserstein distance between the joint feature/labels distribution and a predicted (with the model) joint distribution on the new data. The approach has been recently extended to the deep learning framework in \citep{Damodaran2018} and will be described more in detail in section \ref{sec:jdot}.

\section{Problem formulation and noise model}
\label{sec3}

\subsection{Traditional supervised learning}
\label{sec:}
Let $X =\{\x_1,\ldots, \x_N\} \in \mathcal{X}^N $ be the training features/images and $Y =\{\y_1, \ldots, \y_N\} \in \mathcal{Y}^N$ be their associated one-hot encoded class labels ($\y_{i}\in \{0,1\}^{1\times c}$, $c$ is the number of classes) sampled from the joint distribution $p(\x,\y)$. Let $f: X \rightarrow Y$ be a neural network model with model parameters $\theta$, which maps the input features into class conditional probabilities $f(\x)_j=\hat{p}(\y=j|\x)$. The loss function $L(f(\x),\y)$ measures the discrepancy (error) between the true label $\y$ and the predicted label distribution $f(\x)$ by the neural network. 
In the standard supervised learning stting, one estimates the parameters $\theta$ of $f$ by minimizing the empirical risk on the training set
\begin{equation}
\min_f \quad \frac{1}{N}\sum_{i=1}^{N} L(f(\x_i), \y_i)
\label{eq1:exp_loss}
\end{equation}
In this paper we use the cross-entropy defined as : $L(f(\x_i), \y_i) = \sum_{j=1}^{c} -y_{ij} \log(f(\x_i)_j)$, thus eq. \eqref{eq1:exp_loss} can be re-expressed as
\begin{equation}
\min_f\quad \frac{1}{N}\sum_{i=1}^{N} \sum_{j=1}^{c} -y_{ij} \log(f(\x_i)),
\label{eq:exploss2}
\end{equation}
The  neural network model $f$ is estimated by minimizing the objective above with respect to its parameters through stochastic optimization procedures. However, minimizing the loss function eq.~\eqref{eq:exploss2} in certain scenarios can lead to over-fitting. When the dataset is affected by label noise, minimizing the empirical risk can degrade the performance of the neural network. Hence suitable modification of the loss function is necessary to learn a robust neural network model, which is the direction of our proposed method.

In the following subsection, we describe the label noise in the datasets, and how to artificially simulate this noise in two different settings.

\subsection{Label noise}
Large scale datasets are commonly subjected to label noise (mislabeled samples), especially when using one of the surrogate  labeling strategy discussed in the introduction. The occurrence of the label noise in the dataset can be of two types: asymmetric and symmetric label noise.
 
In the asymmetric label noise, each label $\y$ in the training set is flipped to $\tilde{\y}$ with probability $p(\tilde{\y}|\y)$, defining the noise transition matrix, $E_{i,j}= p(\tilde{\y}=j|\y=i) ~ \forall i, j$, indicating the probability of $i^{th}$ class label being flipped to $j^{th}$ class label.  Thus, the training samples $\{\x_i, \tilde{\y}_i\}$ are observed from the joint distribution
\begin{equation}
p(\x,\tilde{\y}) = \sum_y E_{i,j} p(\y|\x)p(\x),
\end{equation}
 This noise model is realistic and can occur in real world scenario, where non-expert finds difficult to distinguish between the similar fine grained classes. However, the matrix $\mathbf{E}$ is generally unknown in real-world scenarios.

In the symmetric label noise, the label is flipped uniformly across all the classes with probability $p_e$, irrespective of similarity between the classes. In this case, matrix $\mathbf{E}$ has the entries $1-p_e$ in the diagonal, and $\frac{p_e}{1-c}$ in the off-diagonal elements. This noise model is much simpler and has a unique parameter.
  
For both the noise types, learning the classifier $f$ with the loss function mentioned in eq. \eqref{eq1:exp_loss} is not robust and can lead to overfitting to the noisy training labels.

\section{Classification Loss with Entropic Optimal Transport (CLEOT)}
\label{sec4}
In this section we first provide an introduction to optimal transport (OT) by discussing unregularized and regularized OT. Next we introduce the joint distribution OT which is starting point of our method. Then we formulate our approach and discuss the numerical resolution of the proposed learning problem.

\subsection{Introduction to optimal transport}
Optimal transport (see for instance the two monographs by Villani~\citep{villani2003topics, Villani09}) is a theory that allows to compare probability distributions in a geometrically sound manner even when their respective supports do not overlap. OT is hence well-suited to work on empirical distributions and allows to take into account the geometry of the data set in its embedding space.
 Formally, OT searches for a probabilistic coupling $\gamma \in \Pi(\mu_1,\mu_2)$ between two distributions $\mu_1$ and $\mu_2$ which yields a minimal total displacement cost 
wrt. a given cost function $c(\mathbf{x}_1,\mathbf{x}_2)$ measuring the dissimilarity between samples $\mathbf{x}_1$ and $\mathbf{x}_2$ on the support of each distribution $\mu_1$ and $\mu_2$ respectively. Here, $\Pi(\mu_1,\mu_2)$ describes the
space of joint probability distributions with marginals $\mu_1$ and $\mu_2$. In a discrete setting (both distributions are empirical) the OT problem becomes:
 \begin{equation}
 W_C(\mu_1,\mu_2) = \min_{\gamma \in \Pi(\mu_1,\mu_2)}  <\gamma, \mathbf{C}>_F,
 \label{eq:kanto}
 \end{equation}
 where $\langle\cdot, \cdot\rangle_F$ is the Frobenius dot product, $\mathbf{C} \geq 0$ is a ground cost matrix $\in \mathbb{R}^{n_1\times n_2}$ representing the pairwise costs $c(\mathbf{x}_i,\mathbf{x}_j)$, $\gamma$ is a matrix of size $n_1\times n_2$ with prescribed marginals, and $n_1$, $n_2$ the sizes of the supports of the distributions $\mu_1$ and $\mu_2$ respectively. The minimum of this optimization problem can be used as a measure of discrepency between distributions, and, whenever the cost $c$ is a metric, OT is also a metric and is called the Wasserstein distance. 

OT solvers have a super-cubic complexity in the size of the support of the input distributions $n=\max(n_1,n_2)$, which makes OT approaches untractable when dealing with medium to large-scale datasets. In order to speed up OT computation, \cite{Cuturi13} proposed instead of the above linear program to solve a regularized version of OT. Regularization is achieved by adding the negative entropy regularization term to the coupling $\gamma$. Thus, the so-called entropy-regularized Wasserstein distance can be defined as eq.~\eqref{eq:kanto} is becomes
\begin{equation}
 W_{C,\lambda}(\mu_1,\mu_2) = <\gamma^*, \mathbf{C}>_F,
 \label{eq:regot}
\end{equation}
with
\begin{equation}
 \gamma^* = \underset{\gamma \in \Pi(\mu_1,\mu_2)}{\text{argmin}} <\gamma, \mathbf{C}>_F + \lambda R(\gamma)
 \label{eq:regot2}
\end{equation}
where $R(\gamma) = \sum_{i,j} \gamma_{i,j} \log \gamma_{i,j}$ is the negative entropy of $\gamma$, and $\lambda$ is the trade-off between the two terms. When $\lambda=0$ eq. \ref{eq:regot} recovers the original optimal transport problem from eq. \ref{eq:kanto}, and when $\lambda \rightarrow \infty$ the resulting divergence has strong links with maximum mean discrepancy as discussed in \citep{genevay2017sinkhorn}.
Efficient computational schemes were proposed with entropic regularization~\citep{Cuturi13} as well as stochastic versions using the dual formulation of the problem~\citep{genevay2016,arjovsky17,seguy2018}, allowing to tackle middle to large sized problems. 

Note that the regularized Wassersein distance is defined in eq. \ref{eq:regot} only with the linear term whereas the OT matrix $\gamma^*$ is optimized with an additional regularization term in  eq. \ref{eq:regot2}. This allows for a better approximation of the Wasserstein distance as discussed in \cite{luise2018differential}, but comes with a slightly more complex problem to minimize when used as objective value as discussed in the following.

\subsection{Joint distribution optimal transport}
\label{sec:jdot}
In the context of unsupervised domain adaptation,~\cite{courty2017joint} proposed the joint distribution optimal transport (JDOT) method. The idea is to consider the optimal transport problem between distributions on the algebraic product space of features and labels spaces, instead of only considering the feature space distributions.


In this setting, the source measure $\mu_s$ and the target measures $\mu_t$ are measures on the product space $\mathcal{X}\times\mathcal{Y}$, and we note $(x^s, y^s)$, $(x^t, y^t)$ the samples of $\mu_s$ and $\mu_t$ respectively. The generalized ground cost associated to this space can be naturally expressed as a weighted combination of costs in the input and label spaces, reading
 \begin{equation}
d \left( \mathbf{x}_i^s, \mathbf{y}_i^s;\mathbf{x}_j^t, \mathbf{y}_j^t \right) = \alpha c(\mathbf{x}_i^s,\mathbf{x}_j^t) + \beta L(\mathbf{y}_i^s,\mathbf{y}_j^t)
 \label{eq:cost}
 \end{equation}
 for the $i$-th element of the support of $\mu_s$ and $j$-th element of the support of $\mu_t$. $c(\cdot,\cdot)$ is chosen as a $\ell^2_2$ distance and $L(\cdot,\cdot)$ is a classification loss (e.g. hinge or cross-entropy). Parameters $\alpha$ and $ \beta$ are two scalar values weighting the relative contributions of features and label discrepancies. In the unsupervised domain adaptation setting, the labels $\mathbf{y}_j^t$ are unknown and we seek to learn a classifier $f:\mathcal{X}\rightarrow\mathcal{Y}$ to estimate the label $f(\mathbf{x}_j^t)$ of each target sample. Hence, with $(x^t, f(\mathbf{x}_j^t))$ the samples from the target distribution, we define the ground loss,  
 \begin{equation}
    d_f \left( \mathbf{x}_i^s, \mathbf{y}_i^s;\mathbf{x}_j^t \right) = \alpha c(\mathbf{x}_i^s,\mathbf{x}_j^t) + \beta L(\mathbf{y}_i^s,f(\mathbf{x}_j^t))
     \label{eq:cost}
     \end{equation}
 Accounting for the classification loss, JDOT leads to the following minimization problem:
 \begin{equation}
 \min_{f}  W_{\mathbf{D}_f}(\mu_s,\mu_t),
\label{eq:jdot}
 \end{equation}
where  $\mathbf{D}_f$ depends on $f$ and gathers all the pairwise costs $d_f(\cdot,\cdot)$. As a by-product of this optimization problem, samples that share a similar representation and a common label (through classification) are matched, yielding better discrimination. JDOT has been recently extended to deep learning strategies~\citep{Damodaran2018} by computing the optimal transport w.r.t. deep embeddings of the data  rather than the original feature space, and also by proposing a large-scale variant of the regularized OT optimization problem.

\label{sec5}

\subsection{Learning with noisy labels using entropy-regularized OT}

The main idea of our proposed method is to learn a neural network model $f$ efficiently in the presence of noisy labels. 
Let $\{\x_i, \tilde{\y}_i\},~i,~j= 1,\ldots, N$ be the samples and their associated noisy one-hot labels observed from $p(x, \tilde{y})$. We note $\tilde{\mu}$ the discrete distribution corresponding to these samples. Our proposal is to learn $f$ that yields a discrete distribution $\mu_f = \sum_i \delta_{\x_i,f(\x_i)}$ which minimizes the following problem:
 \begin{equation}
 \min_{f}\quad  W_{\mathbf{D}_{f},\lambda}(\tilde{\mu},\mu_f),
\label{eq:eldot}
 \end{equation}
which can be reformulated to the following  bi-level optimization problem:
 \begin{align}
  \min_{f}& \quad \sum_{i,j} T^*_{ij} L\left(\tilde{\y}_i,f(\x_j)\right)
  \label{eq:rselfw}\\
  \text{s.t. }& \gamma^* = \arg\min_\gamma <\gamma, \mathbf{D}_f>_F + \lambda R(\gamma),
  \end{align}
with
\begin{equation}
D_f^{i,j}= \alpha \Vert \x_i - \x_j \Vert^2 + \beta L\left(\tilde{\y}_i,f(\x_j) \right)
\label{eq:jdot_cost}
\end{equation}
As we can see from the objective function eq.~\eqref{eq:rselfw}, $f$ will be learned such that each sample classification $f(\x_j)$ needs to be close to every labels $\tilde{\y_i}$ for which $T_{ij}$ is non-zero. This highlights the role of the optimal coupling $\gamma$ in helping learn a classifier $f$ which is smoother thanks to this averaging process since the OT regularization $\lambda$ will promote a spread of mass in $\gamma$. Here the geometry of the dataset is taken into account through the ground metric on the joint feature-label space.  
This averaging process is even more clear when the classification loss $L$ is linear, which is the case for the cross-entropy loss. Indeed, we have in that case
\begin{equation}
\min_{f} \quad \sum_{i,j} \gamma^*_{ij} L\left(\tilde{\y}_i,f(x_j)\right)=\sum_{j} L\left(\sum_{i}  \gamma^*_{ij}\tilde{\y}_i,f(\x_j)\right)
  \label{eq:label_prop}
 \end{equation}
where  $\sum_{i} \gamma^*_{ij}\tilde{y}_i$ is as an average of the labels with weights in $\gamma$, hence a denoised estimate for $\y_j$. Our approach corresponds to learning from labels that have been smoothed by substituting the each noisy label by a weighted combination of labels where the weights are provided by the optimal couplings estimated w.r.t. the ground cost matrix $\mathbf{D}_{f}$. We name this approach CLEOT for Classification Loss with Entropic Optimal Transport. This approach is notably motivated by the denoising capacity of the entropy regularized optimal problem, as explored in~\citep{Rigollet2018}, and where the denoising is conducted directly on the joint distributions. 
 

In order to better interpret our approach, we can look at the limit cases. When $\beta\rightarrow 0$ and $\lambda=0$ the label loss disappears in the OT metric and the OT matrix is the solution between $\mu$ and itself. In this case the solution is obviously the identity matrix and the optimization problem wrt $f$ boils down to the classical empirical risk minimization of eq. \eqref{eq1:exp_loss} without cross-terms. When  $\lambda=0$ and $\alpha>0,\beta>0$, the OT is performed in the joint distribution sense and will include label information through cost \eqref{eq:jdot_cost}, but the solution will be very sparse (a permutation) at the risk of overfitting the sample assignment in $\gamma^*$. However, when the entropy regularization is included ($\lambda^*>0$), the probability in $\gamma^*$ mass is spread-out, as a result the optimal coupling ($\gamma^*$) will share mass between the samples which have similar features and label representations and perform label smoothing. 
This averaging is of particular interest  when the labels are corrupted by the noise since we always suppose that the good labels are wining locally in average (or else nothing can be learned anyways). Thus, learning the neural network model ($f$) with CLEOT (eq.~\eqref{eq:rselfw}) naturally mitigate the impact of the label noise, and obtains the robust classifier.

Finally we discuss how to solve the optimization problem \eqref{eq:eldot}. The authors of JDOT originally proposed to perform alternative optimization on $\gamma$ and $f$. This approach works for un-regularized OT and converges to a stationary point. However this does not hold true for regularized OT.
When using regularized OT as proposed here, the problem is a bi-level optimization problem \citep{colson2007overview}. Bi-level optimization problem that are notoriously difficult to solve. Since the inner problem is a regularized OT problem that is strongly convex, one could solve the problem by using the implicit function theorem as discussed in \cite{luise2018differential} on a different application. However, solving the full coupling $\gamma^*$ is computationally infeasible both in terms of time and memory, because $\gamma^*$ is a dense matrix and scales quadratically in size to the number of samples. Even if modern solvers have been proposed for regularized OT in the dual \citep{seguy2018,arjovsky17,genevay2016} or primal \citep{genevay2017sinkhorn}, they are still computationally intensive and cannot be used properly with alternate optimization. This problem is even aggravated by the necessity to solve the OT problem at each iteration. 
In order to circumvent these problems, we use a the stochastic optimization scheme by solving the problem on mini-batches, enabling to learn complex deep neural networks on large datasets.   
 
\subsection{Stochastic approximation of proposed method}


We propose to approximate the  objective function eq.~\eqref{eq:rselfw} of our proposed method  by sampling mini-batches of size $m$, and minimizing optimization problem:
\begin{align}
  \min_{f}& \quad \mathbb{E} \left[\quad \sum_{i,j} T^*_{ij} L\left(\tilde{\y}_i,f(\x_j)\right)\right],
  \label{eq:stoch_rselfw}\\
  \text{s.t. }& \gamma^* = \arg\min_\gamma <\gamma, \mathbf{D}_f>_F + \lambda R(\gamma).
  \label{eq:sink_solve}
  \end{align}
where the expectation $\mathbb{E}$ is taken over the randomly sampled mini-batches and \eqref{eq:sink_solve} is solved only on the minibatches. As the size $m$ increases, the optimization problem will converge to eq. \eqref{eq:rselfw}. Still as discussed in~\cite{genevay2017sinkhorn},  the expected value over the mini-batches if OT is not equivalent to the full OT and may lead to a different minimum. In practice it has the effect of densifying the equivalent full OT matrix and adding an additional regularization.  

In order to optimize the problem above on mini-batches we use the \emph{sinkhorn-autodiff} introduced in \citep{genevay2017sinkhorn} that relies on automatic differentiation of the Sinkhorn algorithm that quickly estimates the solution of entropic regularized OT and its gradients (see the psuedo code in algorithm \ref{al:cleot_psuedocode}). Note that we could have used the approach of \cite{luise2018differential} for computing the gradients instead of \emph{sinkhorn-autodiff} but their approach rely on the implicit function theorem which supposes that the inner problem is solved exactly. Since it is difficult to ensure exact convergence of the Sinkhorn, we prefer to perform autodiff on the algorithm with a finite number of iterations which will provide a reasonable gradients even when Sinkhorn has not converged.
This stochastic approach has two major advantages: it scales to large datasets, and can be easily integrated into the modern deep learning framework in an end-to-end fashion.

\begin{algorithm}[!t]
  \begin{algorithmic}[1]
  \Require{
    Training features (images) $\x$, noisy labels $\tilde{\y}$; hyperparameters $\alpha$, $\beta$, $\lambda$}
  \For{each batch ($(\x_1,\tilde{\y}_1),\ldots,(\x_B,\tilde{\y}_B)$)}
   \State Compute the neural network predictions $f(\x_1),\ldots,f(\x_B)$
   \State Compute the OT ground loss as in eq.~\eqref{eq:jdot_cost}
   \State Solve the OT problem ($\gamma^*$) in eq.~\eqref{eq:sink_solve} by Sinkhorn iterations 
   \State Update the neural network parameters by back-propagation
    \EndFor
  \end{algorithmic}
  \caption{Label-noise robust learning}
  \label{al:cleot_psuedocode}
\end{algorithm}


\subsection{Illustration on a toy example}
\begin{figure*}
\begin{subfigure}[l]{0.33\textwidth}
\includegraphics[width=0.8\textwidth]{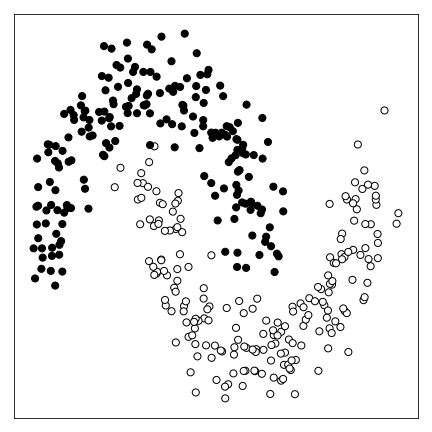}
\caption{Clean dataset}
\end{subfigure}
\begin{subfigure}[l]{0.33\textwidth}
\includegraphics[width=0.8\textwidth]{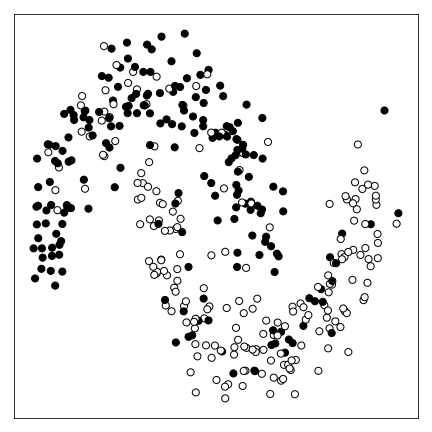}
\caption{noisy version}
\end{subfigure}
\begin{subfigure}[l]{0.33\textwidth}
\includegraphics[width=0.8\textwidth]{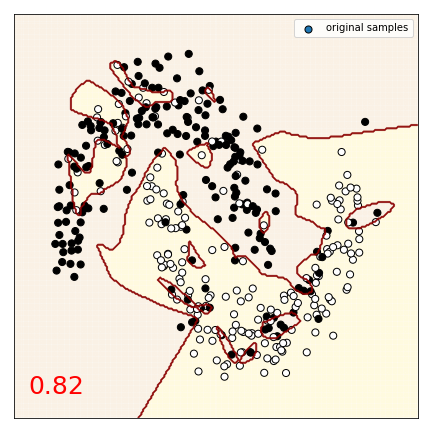}
\caption{Initial decision boundary}
\end{subfigure}\\
\begin{subfigure}[l]{0.33\textwidth}
\includegraphics[width=0.8\textwidth]{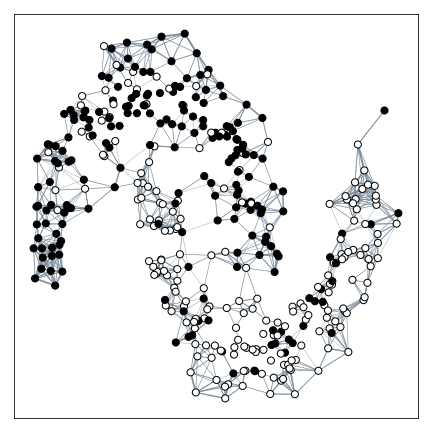}
\end{subfigure}
\begin{subfigure}[l]{0.33\textwidth}
\includegraphics[height=4.9cm,width=\textwidth]{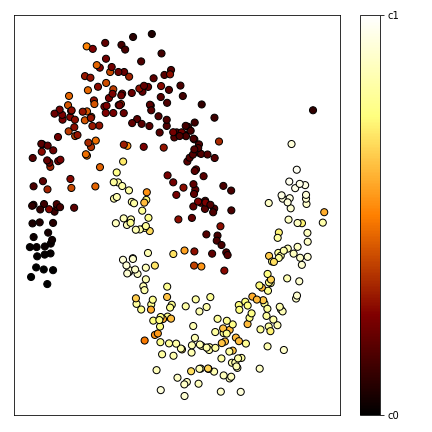}
\end{subfigure}
\begin{subfigure}[l]{0.33\textwidth}
\includegraphics[width=0.8\textwidth]{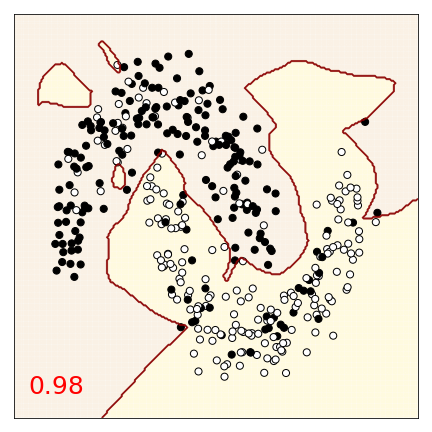}
\end{subfigure}\\
\begin{subfigure}[l]{0.33\textwidth}
\includegraphics[width=0.8\textwidth]{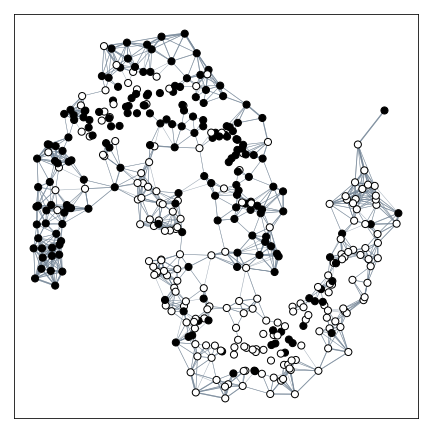}
\end{subfigure}
\begin{subfigure}[l]{0.33\textwidth}
\includegraphics[height=4.9cm,width=\textwidth]{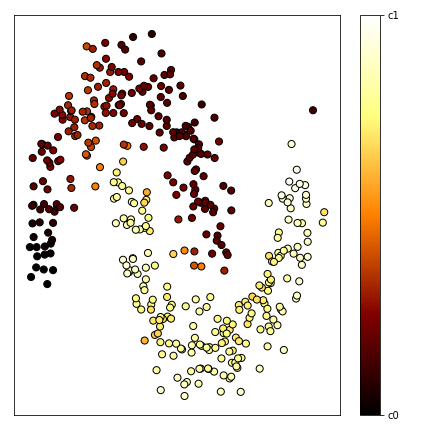}
\end{subfigure}
\begin{subfigure}[l]{0.33\textwidth}
\includegraphics[width=0.8\textwidth]{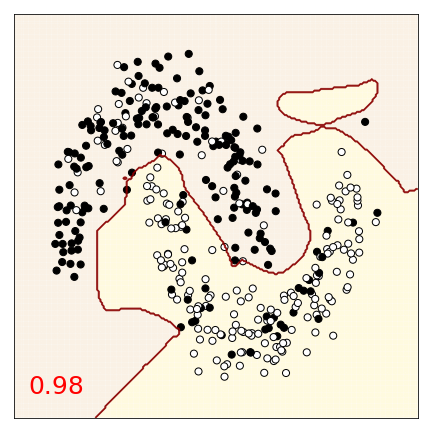}
\end{subfigure}\\
\begin{subfigure}[l]{0.33\textwidth}
\includegraphics[width=0.8\textwidth]{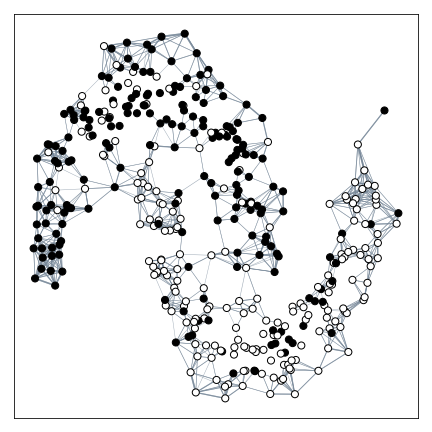}
\caption{Self-coupling}
\end{subfigure}
\begin{subfigure}[l]{0.33\textwidth}
\includegraphics[height=4.9cm,width=\textwidth]{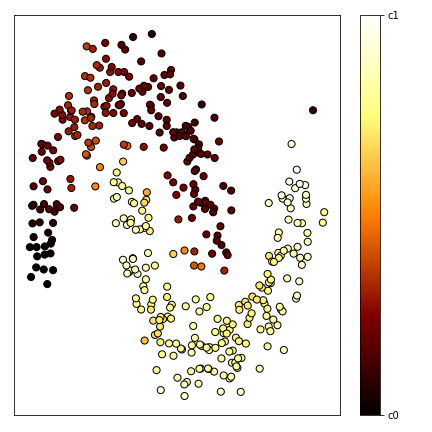}
\caption{Label propagation}
\end{subfigure}
\begin{subfigure}[l]{0.33\textwidth}
\includegraphics[width=0.8\textwidth]{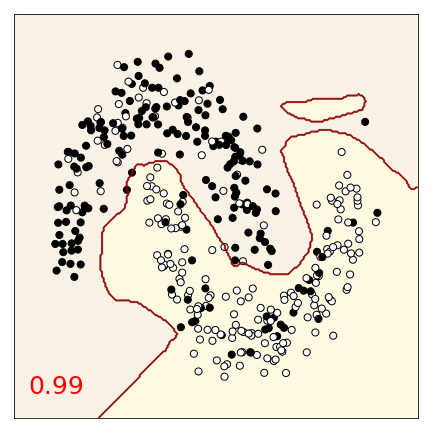}
\caption{Decision boundary}
\end{subfigure}
\caption{Illustration of CLEOT on a toy example along three iterations of the method. First row depicts (a) clean dataset, (b) noisy datasets. Labels are permuted with a probability $p=0.2$ (c) Initial classification by the neural network. Then column (d) shows the self coupling $T$ as a graph (e) label propagation step where each label value is computed as eq.~\eqref{eq:label_prop} (f) new classification boundaries after learning over the propagated labels. On these last images, the classification accuracy is written on the bottom left corner in red.}
\label{fig:toy}
\end{figure*}

For the sake of clarity, we propose an illustration (Figure~\ref{fig:toy}) of the behavior of the method on a simple toy example. It consists in the classical two moons 
problem, which is a binary classification problem. From  a clean version of the dataset (Figure ~\ref{fig:toy}.a), containing 400 data samples, labels are randomly flipped with a probability $p=0.2$ (Figure ~\ref{fig:toy}.b). The classifier is a fully connected neural networks that consists in two hidden layers of size 256, with Relu activations. The model is adjusted along 500 epochs. A graphical representation of the decision boundary is given in (Figure ~\ref{fig:toy}.c), where one can clearly see that the model is not capable of separating properly the two classes, resulting in a complex boundary that encloses mislabeled samples. Then, three iterations of the proposed approach are represented (one per line). Column (d) shows the coupling $\gamma^*$ as a graph, i.e. links between samples corresponds to entries of $\gamma^*$ that are above a given threshold (as $\gamma^*$ is dense). The width of the connection is proportional to the magnitude of the entry. As expected, most of the connections highlight a geometrical and class label proximity. Labels are then propagated (Column (e)) following eq.~\ref{eq:label_prop}. The classifier is  fine tuned over this new set of fuzzy labels. Column (f) shows the new decision boundary, as well as the corresponding accuracy score (in red). As performances increase, it is worth noting  
  the relative lower complexity of the classifier, that almost correctly classifies clean samples (0.99 of accuracy) after three iterations of CLEOT. 
  
\section{Experiments and Results}
\label{sec6}
We evaluate our proposed \texttt{CLEOT} method and state-of-the-art (SoA) methods on two remote sensing tasks: Remote sensing (aerial image) scene classification, and pixel-wise labeling of hyperspectral image. The effectiveness of our proposed method is compared with several SoAs which modifies the loss function similar to ours. The considered SoA methods are \texttt{Backward} and \texttt{Forward} loss correction \citep{Patrini2017}, \texttt{Unhinged} \citep{Rooyen2015}, \texttt{Sigmoid} and \texttt{Ramp} \citep{Ghosh2015}, \texttt{Savage} \citep{Masnadi-ShiraziHamedandVasconcelos2008}, and \texttt{Bootstrap soft} \citep{Reed2015}. The \texttt{Unhinged}, \texttt{Sigmoid}, \texttt{Ramp}, \texttt{Savage} loss correction methods did not perform well in their original form. 
Preliminary experiments show that these methods either do not converge or converge to poor solutions, and sometimes result in premature saturation. In order to make these methods comparable, we stacked the batch-normalization and softmax pooling right before the loss function. This procedure increased the performance of the state-of-the-art methods, compared to the implementation mentioned in \citep{Patrini2017} and in their respective articles. Thus the performance of the  SoA methods can be considered as the strong baseline for our proposed method. A similar procedure is also used for the rest of the methods (including ours) to have uniformity. The source code of our proposed method and SoA methods will be published here once upon acceptance.

In the next subsections,  for each dataset, we first present the data, then detail the label noise simulations and implementation details, and finally present and discuss the results.

\subsection{Aerial Image Labeling}

We have considered four diverse publicly available remote sensing aerial scene classification datasets: \emph{NWPU-RESIS45} \citep{Cheng2017}, \emph{NWPU-19} \citep{Cheng2017}, \emph{PatternNet} \citep{Zhou2018}, \emph{AID} \citep{Xia2017}. The description of each dataset is provided below followed by a description of the label noise applied to them.

\subsubsection{Datasets}
\vspace{1mm}

\paragraph{NWPU-RESIS45} This dataset consists of 31'500 remote sensing images covering 45 scene classes. Each class contains 700 images with a size of 256 $\times$ 256 in the red green blue (RGB) color space. The spatial resolution of this dataset varies from about 30 m to 0.2 m per pixel. This dataset was extracted by the experts in the field of remote sensing image interpretation, from Google Earth (Google Inc.). Additional details of this dataset can be found in \citep{Cheng2017}.

\paragraph{NWPU-19}  This dataset is a subset of NWPU-RESIS45 dateset, which consists of 13'300 remote sensing images divided into 19 scene classes. The number of samples per class, and its size and spatial configuration are similar to NWPU-RESIS45.

\paragraph{PatternNet} This is a large-scale high resolution remote sensing dataset collected for remote sensing image retrieval. Here, we have used it for classification task. It contains 38 classes, and each class consists of 800 images of size 256 $\times$ 256 pixels, totals to 30'400 image scenes. The images in PatternNet are collected from Google Earth imagery or via the Google Map API for US cities. The images are of higher spatial resolution than the NWPU-RESIS45 dataset, the highest spatial resolution is around 0.062 m and lowest is around 4.69 m. For further information, please see \citep{Zhou2018}.

\paragraph{AID} This dataset is made up of 10'000 images covering 30 scene classes. Unlike the above datasets, the number of images in this dataset varies a lot with different aerial scene types, from 220 to 420 sample images. The spatial resolution is varied from 0.5 m to 8 m, and the size of each aerial image is fixed to 600 $\times$ 600 pixels. Similar to above datasets, this dataset is also collected from Google Earth at different time and seasons, over different countries and regions around the world. For more details, please see \citep{Xia2017}.

\subsubsection{Label noise simulation} In order to evaluate our proposed method, we artificially simulate the (asymmetric) label noise in the above datasets, to meet the requirements of real world scenarios. We carefully inspected the samples, and flipped labels according to the noise probability to the visually similar classes. The class permutations that were selected are reported in Table \ref{tab:noise_aerial}.

\begin{table*}\small

  \caption{Table of the selected label noise on the Aerial Image Labeling datasets. The classes which are flipped according to the label noise is described below. $a\rightarrow  b$ indicates the label $a$ is flipped into class label $b$, and  $a\leftrightarrow  b$ indicates that the labels are flipped in both direction. \label{tab:noise_aerial}}

  \begin{tabular}{|p{.18\textwidth}|p{.77\textwidth}|} \hline
    Dataset & Label noise\\
    \hline\hline
    \emph{NWPU-RESIS45} 14/45 classes impacted 
    &baseball diamond $\rightarrow$ medium residential, beach $\rightarrow$ river, dense residential $\leftrightarrow$ medium residential, intersection $\rightarrow$ freeway, mobile home park $\leftrightarrow$ dense residential, overpass $\leftrightarrow$ intersection, tennis court $\rightarrow$ medium residential, runway $\rightarrow$ freeway, thermal power station $\rightarrow$ cloud,
    wetland $\rightarrow$ lake, rectangular farm land $\rightarrow$ meadow, church $\rightarrow$ palace, commercial area $\rightarrow$ dense residential \\ \hline
    \emph{NWPU-19}  \hspace{2cm}7/19 classes impacted
    &baseball diamond $\rightarrow$ medium residential,  beach $\rightarrow$ river, dense residential $\leftrightarrow$ medium residential, intersection $\leftrightarrow$ freeway, mobile home park $\leftrightarrow$ dense residential, overpass $\leftrightarrow$ intersection, tennis court $\rightarrow$ medium residential. \\ \hline
    \emph{PatternNet} \hspace{2cm} 11/38 classes impacted
    &cemetery $\rightarrow$ christmas tree farm, harbor $\leftrightarrow$ ferry terminal, dense residential $\rightarrow$ coastal home, overpass $\leftrightarrow$ intersection, parking space $\rightarrow$ parking lot, runway mark $\rightarrow$ parking space, coastal home $\leftrightarrow$ sparse residential, swimming pool $\rightarrow$ coastal home \\\hline
    \emph{AID} \hspace{2cm}  12/30 classes impacted
    &bareland $\rightarrow$ desert, centre $\rightarrow$ storage tank, church $\rightarrow$ centre, storage tank; dense residential $\rightarrow$ medium residential, desert $\rightarrow$ bareland, industrial $\rightarrow$ medium residential, meadow $\rightarrow$ farm land, medium residential $\rightarrow$ dense residential, play ground $\rightarrow$ meadow, school; resort $\rightarrow$ medium residential, school $\rightarrow$ medium residential, play ground; stadium $\rightarrow$ play ground \\\hline
  \end{tabular}

\end{table*}

\subsubsection{Model} We employed pre-trained VGG16 architecture, replacing the last layer with two MLPs that map to 512 hidden neurons before predicting the classes with $l_2=1e^{-3}$ regularization, respectively. The dropout layer with $p=0.5$ is inserted before the last MLP and the batch normalization is applied before the softmax operator. To have uniformity, all the methods follow similar architecture design.  During the training, the network is fine-tuned by freezing the weights of the VGG16 layers. We optimized the SoA methods for $300$ epochs using the SGD optimizer ($lr=0.01$) with momentum ($m=0.9$) using the mini-batch size of 128. The proposed \texttt{CLEOT} method is also optimized as above, but with different learning rate ($lr=0.1$) and mini-batch size (50 samples per class). The hyper-parameters of \texttt{CLEOT} method are set as $\alpha=1$, $\beta=0.005$, and $\lambda=0.005$ experimentally for all the datasets. Additionally, we have used early stopping criterion to terminate the training process, if the validation loss did not decrease for $25$ epochs. This allows to prevent over-fitting to the noisy labels for all the methods. Furthermore, we retained the model weights with best validation loss.

For all the datasets, from the available number of samples we partitioned 80\% of samples for training, 10\% samples for validation and the remaining 10\% samples for evaluating the performance. All the methods are trained with the noisy labeled training and validation samples, and evaluated with the clean testing label instances.

\subsubsection{Results}
\begin{table*}[ht]
\caption{The average classification accuracies and standard deviation of SoA methods and proposed \texttt{CLEOT} method on remote sensing aerial scene classification datasets. The accuracy measures are averaged over 5 runs. and the best accuracies are reported in \textbf{bold}.}
\resizebox{\textwidth}{!}{
\begin{tabular}{l|lllll|lllll}
\hline
Method &\multicolumn{5}{|c|}{NWPU-45} & \multicolumn{5}{|c}{NWPU-19}\\
\centering
  & 0.0 &0.2 &0.4 &0.6 &0.8 & 0.0 &0.2 &0.4 &0.6&0.8 \\
  \hline \hline
Cross entropy &82.93$\pm$0.09& 80.53$\pm$0.30& 75.63$\pm$1.04& 67.80$\pm$0.40&62.05$\pm$0.04

 & 89.75$\pm$0.30& 84.95$\pm$0.49& 78.19$\pm$2.08& 68.45$\pm$0.95 &62.25$\pm$0.17 \\

Unhinged &82.81$\pm$0.21&{82.13$\pm$0.14}& 78.38$\pm$0.57& 63.07$\pm$0.26
&61.01$\pm$0.01
& 90.07$\pm$0.20& 87.36$\pm$0.46& 79.22$\pm$0.77& 66.65$\pm$0.85 &61.22$\pm$0.41 \\

Sigmoid &71.74$\pm$0.40& 68.08$\pm$0.18& 65.76$\pm$0.50& 57.10$\pm$0.06&56.61$\pm$0.31

&
89.69$\pm$0.06&  88.73$\pm$0.19 & {84.37$\pm$0.09}& 66.19$\pm$0.29&59.62$\pm$0.11\\

Ramp &82.99$\pm$0.10& \textbf{82.26$\pm$0.20}& {78.81$\pm$0.26}& 62.97$\pm$0.16 &60.91$\pm$0.32
&
\textbf{90.77$\pm$0.28}& 86.69$\pm$0.18& 78.62$\pm$0.77&67.25$\pm$0.60&60.70$\pm$0.44 \\

Savage &76.85$\pm$0.15& 75.13$\pm$0.11& 69.96$\pm$0.14& 59.56$\pm$0.03&58.08$\pm$0.07 

&
{90.20$\pm$0.18}& {89.01$\pm$2.80} & 81.13$\pm$0.41 & 67.21$\pm$0.20 & 60.62$\pm$0.12\\

Bootstrap soft &82.98$\pm$0.17& 80.65$\pm$0.47& 75.82$\pm$0.88& 67.39$\pm$0.86 &62.22$\pm$0.21

&
89.74$\pm$0.18& 85.19$\pm$0.56& 79.64$\pm$0.95&{69.20$\pm$1.50}&62.03$\pm$0.05 \\

Backward $\hat{E}$  &82.79$\pm$0.14& 80.65$\pm$0.51& 75.96$\pm$0.72& {68.67$\pm$0.75 }&62.45$\pm$0.52
&89.73$\pm$0.43& 85.20$\pm$0.37& 78.17$\pm$1.01&68.72$\pm$1.60&62.06$\pm$0.08\\

Forward $\hat{E}$  &\textbf{83.06$\pm$0.11}& 80.87$\pm$0.53& 74.97$\pm$1.02& 68.12$\pm$1.16 &62.56$\pm$0.16 
& 89.97$\pm$0.32 & 85.37$\pm$1.04& 78.89$\pm$1.28 &69.07$\pm$0.95 & 62.31$\pm$0.24\\\hline

CLEOT &82.41$\pm$0.27& \textbf{81.54$\pm$0.18}& \textbf{80.84$\pm$0.45}& \textbf{76.07$\pm$0.35} & \textbf{70.14$\pm$0.33}&
 89.98$\pm$0.38& \textbf{89.17$\pm$0.30}& \textbf{86.26$\pm$0.63}&\textbf{78.04$\pm$0.42} &\textbf{68.08 $\pm$ 0.53}\\
\hline
\hline
 &\multicolumn{5}{|c|}{PatternNet} & \multicolumn{5}{|c}{AID}\\
  & 0.0 &0.2 &0.4 &0.6&0.8  & 0.0 &0.2 &0.4 &0.6 &0.8\\
  \hline \hline
Cross entropy & 97.68$\pm$0.15& 94.82$\pm$0.30& 89.11$\pm$0.55& 79.76$\pm$1.36 &
73.68$\pm$0.34
& 86.94$\pm$0.51& 82.92$\pm$0.33& 73.80$\pm$1.03& 65.56$\pm$1.18&57.80$\pm$0.84 \\

Unhinged &\textbf{97.76$\pm$0.16} & {97.55$\pm$0.05}&{95.31$\pm$0.19}&73.62$\pm$0.59 &71.23$\pm$0.14
& {87.64$\pm$0.19} & \textbf{86.33$\pm$0.19} & 78.67$\pm$0.29 & 65.93$\pm$0.52 &57.20$\pm$0.16 \\

Sigmoid &96.63$\pm$0.03 & 96.31$\pm$0.31&94.63$\pm$0.24&73.17$\pm$0.39 & 70.58$\pm$0.04
& 85.41$\pm$0.26 & 84.71$\pm$0.25 & {82.05 $\pm$0.17} & 60.96$\pm$0.44 &56.18$\pm$0.08\\

Ramp &{97.73$\pm$0.07} & \textbf{97.56$\pm$0.02}&\textbf{95.44$\pm$0.16}&72.94$\pm$0.47 &71.37$\pm$0.16
& \textbf{87.74$\pm$0.22} & {86.24$\pm$0.23} &78.37$\pm$0.56 & {66.04$\pm$0.59} &57.21$\pm$0.27 \\

Savage  &96.82$\pm$0.05 & 96.41$\pm$0.03&93.94$\pm$0.11&73.16$\pm$0.13 &70.72$\pm$0.01

& 83.65$\pm$0.10 &85.73$\pm$0.21 & \textbf{82.28$\pm$0.27} & 62.88$\pm$0.50&56.55$\pm$0.48 \\

Bootstrap soft  &97.62$\pm$0.13& 94.45$\pm$0.39& 88.88$\pm$0.79& 79.13$\pm$0.73 &73.39$\pm$0.48
& 87.03$\pm$0.40& 82.54$\pm$0.78& 73.75$\pm$0.82& 65.24$\pm$1.09&58.00$\pm$0.45\\

Backward $\hat{E}$  &97.60$\pm$0.10& 94.76$\pm$0.31& 89.07$\pm$0.70& {79.89$\pm$0.36} &73.47$\pm$0.32 
&86.87$\pm$0.52& 82.63$\pm$0.59& 74.03$\pm$0.56& 65.71$\pm$1.16 & 57.90$\pm$0.22\\

Forward $\hat{E}$  &97.67$\pm$0.06& 94.43$\pm$0.78& 89.16$\pm$1.01& 79.44$\pm$0.51 &
73.21$\pm$0.62
&86.91$\pm$0.41& 82.30$\pm$1.08& 73.59$\pm$0.76& 64.91$\pm$0.74 &58.43$\pm$0.59\\\hline

CLEOT &97.29$\pm$0.04& 96.77$\pm$0.09& 94.51$\pm$0.15& \textbf{83.75$\pm$0.20}& \textbf{79.84 $\pm$ 0.22}
 & 87.02$\pm$0.63 & 85.39$\pm$1.12 &79.19$\pm$0.94 & \textbf{71.76$\pm$0.66} &\textbf{63.23$\pm$0.42}\\
\hline
\hline
\label{tab:aerial}
\end{tabular}
}
\end{table*}

Table \ref{tab:aerial} presents the classification performance of our proposed and SoA methods on the four aerial scene classification datasets with different noise levels. We have also included \texttt{Cross entropy} loss function, which is the baseline for all the approaches. The noise level $0$ indicates that the methods are trained with the clean labeled training and validation samples, and it can be considered as the gold standard. The impact of label noise is varied and analyzed in the range of $p_e=\{ 0, 0.2, 0.4, 0.6, 0.8\}$. The amount of actual noise depends on the number of the classes affected by the label noise in the dataset.

When the conventional \texttt{Cross entropy} loss function is considered, the classification accuracy drops to few percentage of points (3-5\%) initially and decreases drastically (above 15\%)  as the magnitude of label noise increases with all the four datasets. This shows that regularization techniques such as weight regualizers, dropout and early stopping criteria can circumvents  label noise only up to some degree and is inefficient in high level label noise. This emphasizes need for the inclusion of robust loss functions while training the deep neural networks in remote sensing image analysis. Next, when the performance of existing SoA methods are considered, they showed robust performance and did not degrade the performance compared to the clean training set under the noise level $0.2$ but decreases about 4\% on the mid noise level, however they outperformed the conventional \texttt{Cross entropy} loss function.  Further, it is noted that under the high level label noise, the SoA methods are similar or less than the \texttt{Cross entropy} loss function. Thus, SoA methods are still limited to tackle the complex noise scenarios.

Lastly, when the performance of the proposed \texttt{CLEOT} method is analyzed, one could see that \texttt{CLEOT} achieves better or similar performance to the SoA methods in the low-level noise, and achieves impressive performance on the higher noise levels. For instance, on average our method decreases only 2.6\%, 10.6\% compared to clean training set with 0.4, and 0.6 noise level, where as the best SoA decreases about 4\%, and 22.5\% respectively. Further, \texttt{Forward} and \texttt{Backward} methods are inferior to the the robust loss functions (\texttt{Unhinged, Sigmoid}, etc), which is contrary w.r.t to the observation in \cite{Patrini2017}. This reveals that methods that perform well on machine learning datasets not necessarily achieve better performance in remote sensing datasets, thus new methods has to be designed specific to remote sensing datasets. 
Lastly, it is noted from table \ref{tab:aerial} that, among the SoA methods there is no single best method which consistently performs better across the datasets, and noise levels. 
Thus, there exists dilemma in choice of method among the existing SoA for the underlying real world task. On contrary, our method consistently outperforms across different datasets, and noise levels.

\subsection{Hyperspectral image classification}
Next, we evaluate our proposed method on the pixel-wise labeling task of hyperspectral datasets. For this, we have chosen three hyperspectral datasets from three different type of sensors covering agricultural and urban cover settings.
\subsubsection{Datasets}

\paragraph{Pavia University} The first hyperspectral data considered here was collected over the University of Pavia,
Italy by the ROSIS airborne hyperspectral sensor in the
framework of the HySens project managed by DLR
(German national aerospace agency). The ROSIS sensor
collects images in 115 spectral bands in the spectral
range from 0.43 to 0.86 µm with a spatial resolution
of 1.3 m/pixel. After the removal of noisy bands, 103
bands were selected for experiments. This data contains
610$\times$340 pixels with nine classes of interest, which covers the urban materials. The total number of available labeled ground truth samples is 42'776.

\paragraph{Chikusei} The airborne hyperspectral dataset was taken by Headwall Hyperspec-VNIR-C imaging sensor over agricultural and urban areas in Chikusei, Ibaraki, Japan. The hyperspectral dataset has 128 bands in the spectral range from 363 nm to 1018 nm. The scene consists of 2517$\times$2335 pixels and the ground sampling distance was 2.5 m. Ground truth of 19 classes was collected via a field survey and visual inspection using high-resolution color images obtained by Canon EOS 5D Mark II together with the hyperspectral data. The number of labeled reference samples is 77'592. For additional details, please refer~\citep{NYokoya2016}.
 
\paragraph{GRSS\_DFC\_2018} The last hyperspectral dataset used were acquired over the University of Houston campus and its neighborhood  on February 2017 by an ITRES CASI 1500 imaging sensor. This dataset contains 48 spectral bands covering the spectral range of 380 nm to 1050 nm with 1 m ground sampling distance. The scene consists of 601$\times$2384 pixels representing 20 urban land use/cover classes, and contains 50'4856 labeled reference samples. The details of this dataset can be found in \footnote{http://www.grss-ieee.org/community/technical-committees/data-fusion}.

\begin{table*}\small

  \caption{Table of the selected label noise on the Hyperspectral image classification datasets. The classes which are flipped according to the label noise is described below. $a\rightarrow  b$ indicates the label $a$ is flipped into class label $b$, and  $a\leftrightarrow  b$ indicates that the labels are flipped in both direction. \label{tab:noise_hyper}}

  \begin{tabular}{|p{.18\textwidth}|p{.77\textwidth}|} \hline
    Dataset & Label noise\\
    \hline\hline
    \emph{Pavia University}
    
    7/9 classes impacted 
    &meadows $\leftrightarrow$ trees,
    gravel $\leftrightarrow$ self building blocks, bare soil $\rightarrow$ meadows, bitumen $\leftrightarrow$ asphalt. Out of 9 classes, 7 classes are impacted by the label noise. \\ \hline
    \emph{Chikusei}  \hspace{2cm}10/19 classes impacted
    &baresoil (park) $\rightarrow$ baresoil (farm); baresoil (farm) $\rightarrow$ baresoil (park), rowcrops;
    weeds $\rightarrow$ grass, rowcrops, forest; forest $\rightarrow$ rice(grown), weeds; grass $\rightarrow$ weeds, rowcrops; rice(grown) $\rightarrow$ forest, weeds; rowcrops $\rightarrow$ baresoil(farm), weeds, grass; plastic home $\rightarrow$ asphalt, manmade(dark); manmade(dark) $\rightarrow$ plastic house;paved ground $\rightarrow$ baresoil(farm) \\ \hline
    \emph{GRSS\_DFC\_2018} \hspace{2cm} 10/20 classes impacted
    &healthy grass $\rightarrow$ stressed grass; stressed grass $\rightarrow$  bare earth; evergreen trees  $\rightarrow$  deciduous trees; deciduous trees  $\rightarrow$  residential buildings; residential buildings  $\rightarrow$  roads, sidewalks;  non-residential buildings  $\rightarrow$  sidewalks; roads $\rightarrow$ major thoroughfares, sidewalks; sidewalks $\rightarrow$  major thoroughfares, crosswalks; crosswalks $\rightarrow$ major thoroughfares; major thoroughfares $\rightarrow$ highways \\\hline
  
  \end{tabular}

\end{table*}

\subsubsection{Label noise simulation}
For the hyperspectral datasets, it is difficult to find similar classes with visual inspection due to the high dimensionality of the data. So we measure class similarity using the Jeffries-Matusita distance and Transformed divergence measure \citep{Richards1999} and visual interpret the spectral signatures of some training samples for the most similar classes. According, the class labels are flipped as defined in Table \ref{tab:noise_hyper}.

\subsubsection{Model}

We used a recent state-of-the-art hyperspectral image classification framework named spectral-spatial residual residual network (SSRN) \citep{Zhong2018}. It consecutively extracts spectral and spatial features for pixel wise classification of hyperspectral image. The spectral feature learning consists of two 3-D convolutional layer, and two residual blocks. Following the spectral features, spatial features are extracted using 3-D convolutional layer and two spatial residual blocks. Average pooling layer is added on top of the spectral-spatial feature volume, and followed by the fully connected layer with softmax activation function. Dropout layer (p=0.5) is added after the average pooling layer and  batchnormalization layer is stacked before the softmax activation. Please refer to \citep{Zhong2018} for additional details of the SSRN architecture. 

We trained the SSRN architecture using SGD momentum optimizer with lr = 0.01 and m = 0.9 for 600 epochs using the batch size of 128 for the SoA methods, and 256 for the \texttt{CLEOT}.  As with aerial scene classification we also employed the eary stopping criterion to avoid overfitting, and terminate the training process, if the validation loss did not decrease for $15$ epochs. All the methods follow similar training procedure. The hyperparameters of our proposed method \texttt{CLEOT}  are set to $\alpha=1$, $\beta=0.05$ experimentally for all the datasets, and the entropic regularizer is set to $\lambda=0.05$ for the PaviaU and Chikusei datasets, and $\lambda=1$ for the remaining dataset. 

While partitioning the ground truth reference samples into training, validation and testing subsets, we followed the conventional protocol in the hyperspectral remote sensing community to train classifier with small number of training samples. Accordingly, we used 20\% of samples for training, 10\% samples for validation and remaining 70\% samples for testing purpose for the Pavia University, and Chikusei datasets. Where as for the GRSS\_DFC\_2018 dataset, we used 10\% samples for training, 10\% samples for validation and remaining 80\% samples for evaluation. The training and validation samples are impacted by the label noise, and clean testing labeled samples is used for evaluation.

\subsubsection{Results}

Table \ref{tab:hsi_result} presents the classification performance of SoA methods and \texttt{CLEOT} for the three hyperspectral datasets. The experiments are conducted with different noise levels (see Table \ref{tab:hsi_result} for noise levels) to effectively access the robustness of SoA and proposed methods.
The noise level $p_e=0$ indicates the clean labeled training and validation samples, which is an upper bound for all the methods.

\begin{table*}[htbp]
  \caption{The average classification accuracies and standard deviation of SoA methods and our proposed \texttt{CLEOT} method on the pixel-wise labeling task of hyperspectral datasets. The accuracy measures are computed over three runs, and the best accuracies are reported in \textbf{bold}.}
  \resizebox{\textwidth}{!}{
  \begin{tabular}{l|lllll|llllll}
  \hline
  Method &\multicolumn{5}{|c|}{PaviaU} & \multicolumn{6}{|c}{GRSS\_DFC\_2018}\\
   & 0.0 &0.1 &0.2&0.3 &0.4 & 0.0 &0.1 &0.2 &0.3&0.4&0.5 \\
   & (0.0) & (0.19) & (0.37)& (0.57)& \\
    \hline \hline
    Cross entropy & 99.93$\pm$0.02&95.62$\pm$0.38 &85.31$\pm$0.60&78.01$\pm$1.78 &65.13$\pm$1.49  
  & 84.93$\pm$2.27 &68.66$\pm$16.13& 76.71$\pm$6.37&79.54$\pm$16.57&71.54$\pm$14.54 &44.33$\pm$11.52    \\

  Unhinged  & 99.97$\pm$0.02 & 98.58$\pm$0.01 &94.91$\pm$0.01 &93.02$\pm$0.02&78.55$\pm$0.01 
   & 87.07$\pm$1.48&94.23$\pm$1.03&85.24$\pm$3.85 &92.76$\pm$0.65&75.90$\pm$2.21 &45.11$\pm$4.45 
   \\
  
  Sigmoid &\textbf{99.98$\pm$0.00} &\textbf{99.68$\pm$0.05} &97.56$\pm$0.01 &88.69$\pm$0.01&69.31$\pm$0.02
   & 90.33$\pm$6.72&84.87$\pm$7.01&90.18$\pm$4.09 &86.23$\pm$3.79&85.18$\pm$5.31 &54.04$\pm$6.11
   \\
  
  Ramp & 99.59$\pm$0.49& 98.58$\pm$0.17 &96.04$\pm$0.65 &90.70$\pm$2.02&75.75$\pm$4.65
  & 92.32$\pm$3.11&85.91$\pm$8.29 &77.50$\pm$3.32&85.70$\pm$3.69&82.01$\pm$8.13&40.19$\pm$8.92
   \\
  
  Savage & 99.97$\pm$0.01& 95.91$\pm$0.86 &86.07$\pm$0.38&76.06$\pm$0.08&64.84$\pm$1.33 
  & 83.09$\pm$10.68&92.20$\pm$3.27 &88.42$\pm$4.15&91.69$\pm$1.43&83.96$\pm$3.50 &48.23$\pm$15.16
   \\
  
  Bootstrap soft & 99.89$\pm$0.01& 91.44$\pm$3.28 &85.96$\pm$2.34 &75.27$\pm$1.18&66.07$\pm$2.25
  & 85.57$\pm$13.96&87.55$\pm$5.59 &87.93$\pm$5.96 &80.07$\pm$3.81&75.92$\pm$12.42 &28.37$\pm$7.38
   \\
  
  Backward $\hat{E}$  &99.94$\pm$0.02&90.91$\pm$3.52 &84.08$\pm$0.64&76.26$\pm$3.39 &71.65$\pm$6.74 
  &85.51$\pm$7.97 &69.67$\pm$25.91 &84.44$\pm$5.94&86.08$\pm$4.64 &78.40$\pm$11.54 &71.13$\pm$12.40
   \\
  
  Forward $\hat{E}$  & 96.65$\pm$4.65& 95.74$\pm$0.26 &87.64$\pm$1.05&84.05$\pm$3.24&65.46$\pm$5.09 
  & 79.26$\pm$9.11&88.50$\pm$0.50 &88.84$\pm$3.04&87.35$\pm$6.44&85.44$\pm$5.28& \textbf{83.30$\pm$4.46} 
   \\\hline
  
  CLEOT &99.80$\pm$0.12 &99.28$\pm$0.24 &\textbf{98.42$\pm$0.21} & \textbf{96.91$\pm$0.50} & \textbf{91.31$\pm$0.23} 
  
  &\textbf{96.01$\pm$0.59}& \textbf{95.50$\pm$0.56} & \textbf{94.81$\pm$0.50} & \textbf{92.18$\pm$1.51} & \textbf{91.08$\pm$0.15} & 62.98$\pm$1.65
    \\
  \hline
  \hline
  
   \hline 
  Method &\multicolumn{7}{|c}{chikusei} \\
   & 0.0 &0.1 &0.2 &0.3  & 0.4 &0.5 &0.6 \\
    \hline \hline
    Cross entropy  &\textbf{99.99$\pm$0.01} &99.27$\pm$0.04&98.98$\pm$0.02&95.92$\pm$0.17&92.40$\pm$0.42&86.37$\pm$0.31&61.31$\pm$0.26    \\
  
  Unhinged  & 99.87$\pm$0.18 & 99.74$\pm$0.16&99.45$\pm$0.09&99.19$\pm$0.01&97.57$\pm$0.22&91.72$\pm$0.37&67.29$\pm$7.45 
     \\
  
  Sigmoid  &99.44$\pm$0.01 & 99.51$\pm$0.01 &99.42$\pm$0.01 &99.23$\pm$0.02 &99.06$\pm$0.02 & 94.35$\pm$0.06&64.02$\pm$0.03  
     \\
  
  Ramp &99.88$\pm$0.15 & 99.16$\pm$0.64 &99.50$\pm$0.01 &99.17$\pm$0.13 &97.69$\pm$0.75 & 91.81$\pm$0.96&64.79$\pm$7.15  
    \\
  
  Savage &99.99$\pm$0.02 & \textbf{99.98$\pm$0.00} &98.67$\pm$0.01 &90.78$\pm$1.45 &83.39$\pm$3.81 &66.87$\pm$2.46&52.40$\pm$0.05 
   \\
  
  Bootstrap soft &99.95$\pm$0.04 & 99.59$\pm$0.16 &98.73$\pm$0.31 & 95.79$\pm$1.44 &92.35$\pm$0.48 & 81.39$\pm$7.73&63.48$\pm$4.44 
     \\
  
  Backward $\hat{E}$  &99.92$\pm$0.10& 97.74$\pm$1.43&97.34$\pm$1.24&95.81$\pm$0.90&89.68$\pm$5.60&79.26$\pm$4.32&65.31$\pm$9.60
    \\
  
  Forward $\hat{E}$ &99.99$\pm$0.01& 99.17$\pm$0.01&98.98$\pm$0.04&96.84$\pm$0.05&95.58$\pm$0.06&83.89$\pm$0.06&64.15$\pm$0.04 
     \\\hline
  
  CLEOT &99.88$\pm$0.01 &99.41$\pm$0.13 & \textbf{99.59$\pm$0.12} & \textbf{99.24$\pm$0.01} & \textbf{99.10$\pm$0.02} & \textbf{96.64$\pm$1.20} & \textbf{84.50$\pm$1.35}
     \\\hline
     \hline
  \end{tabular}
  }
  \label{tab:hsi_result}
  \end{table*}

\begin{figure*}[ht]
\begin{subfigure}[l]{0.45\textwidth}
\includegraphics[scale=0.5]{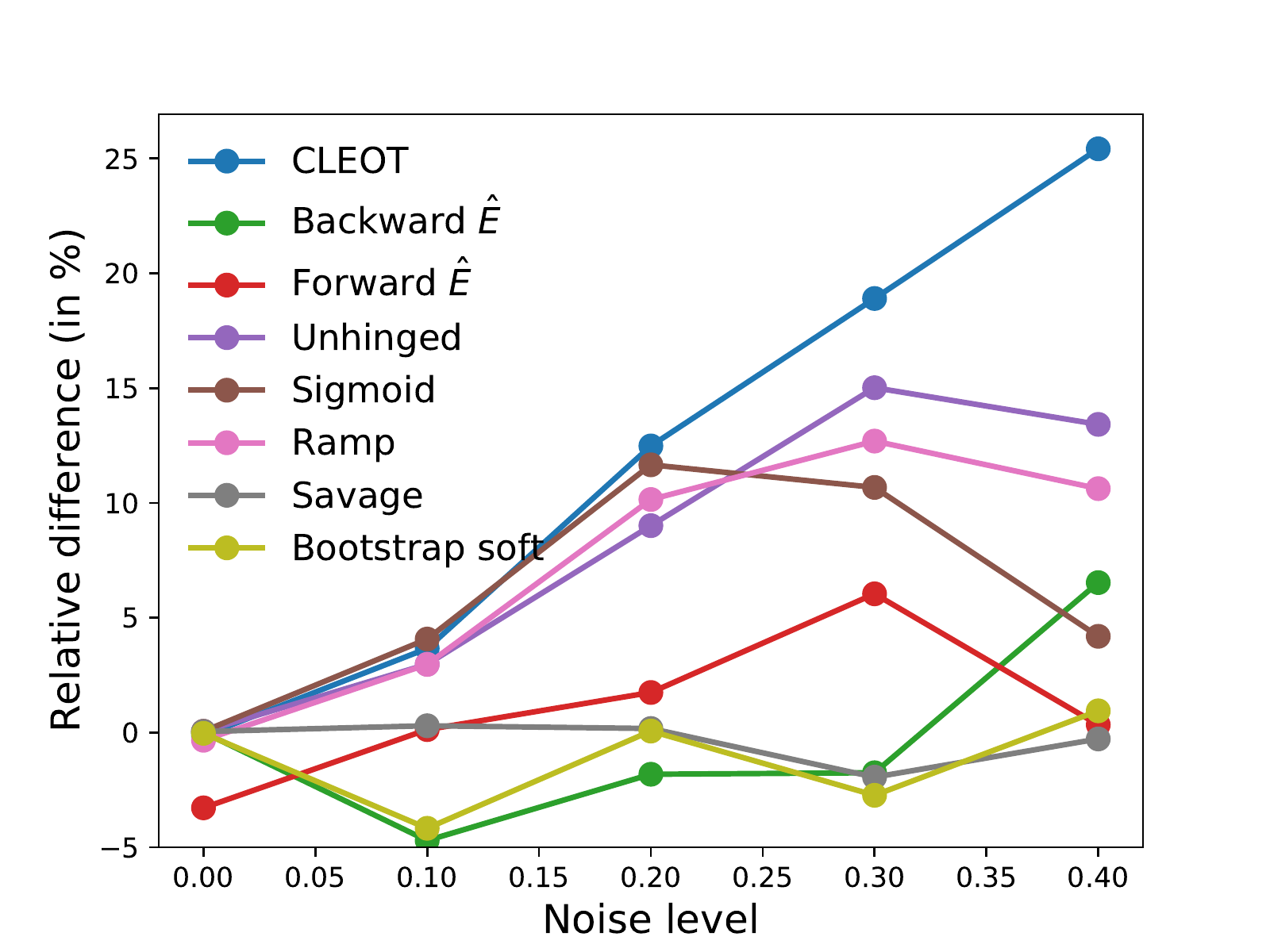}
\caption{Pavia University}
\end{subfigure}
\begin{subfigure}[l]{0.45\textwidth}
\includegraphics[scale=0.5]{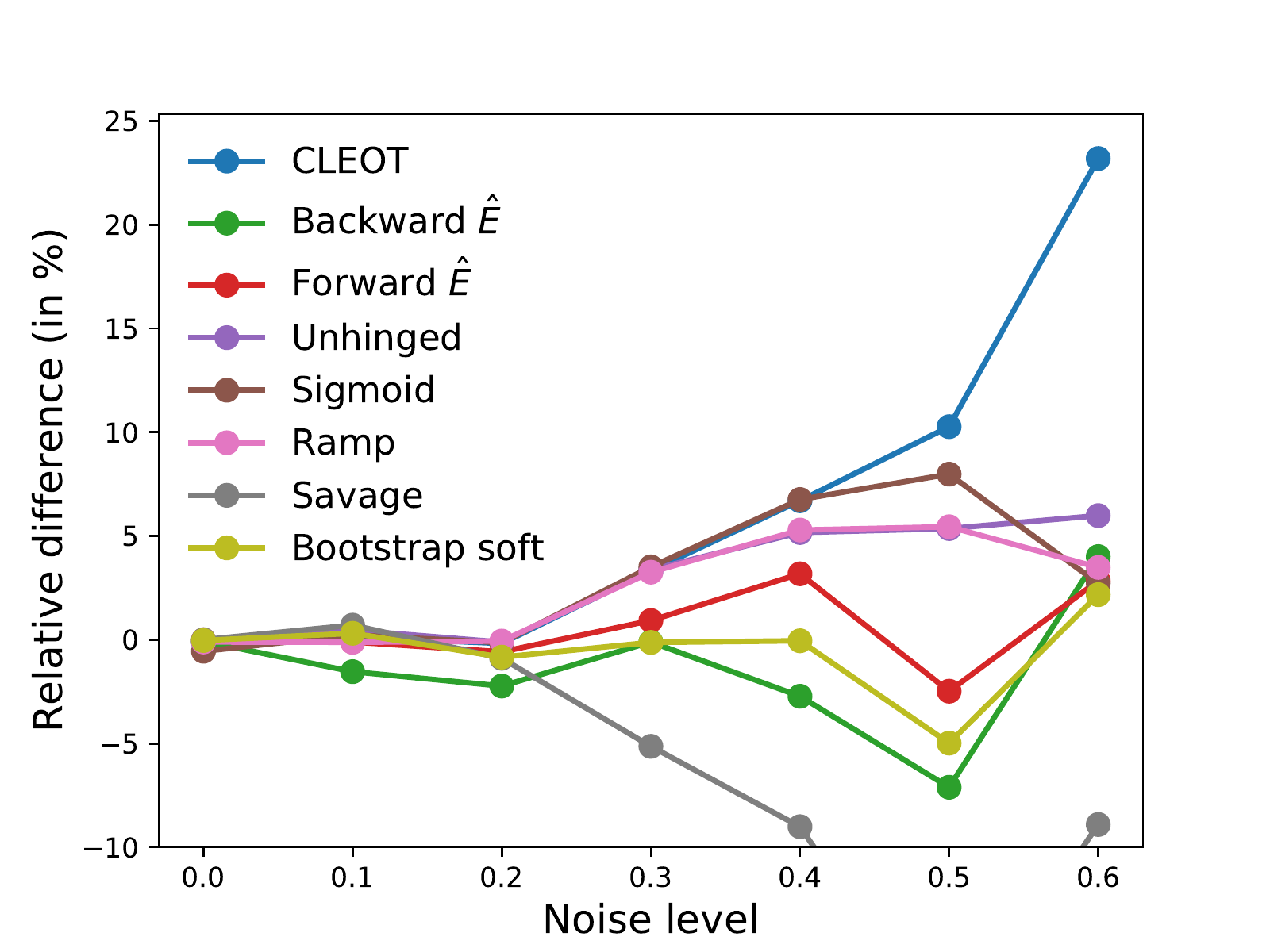}
\caption{Chikusei}
\end{subfigure}
\begin{subfigure}[c]{\textwidth}
\centering
\includegraphics[scale=0.5]{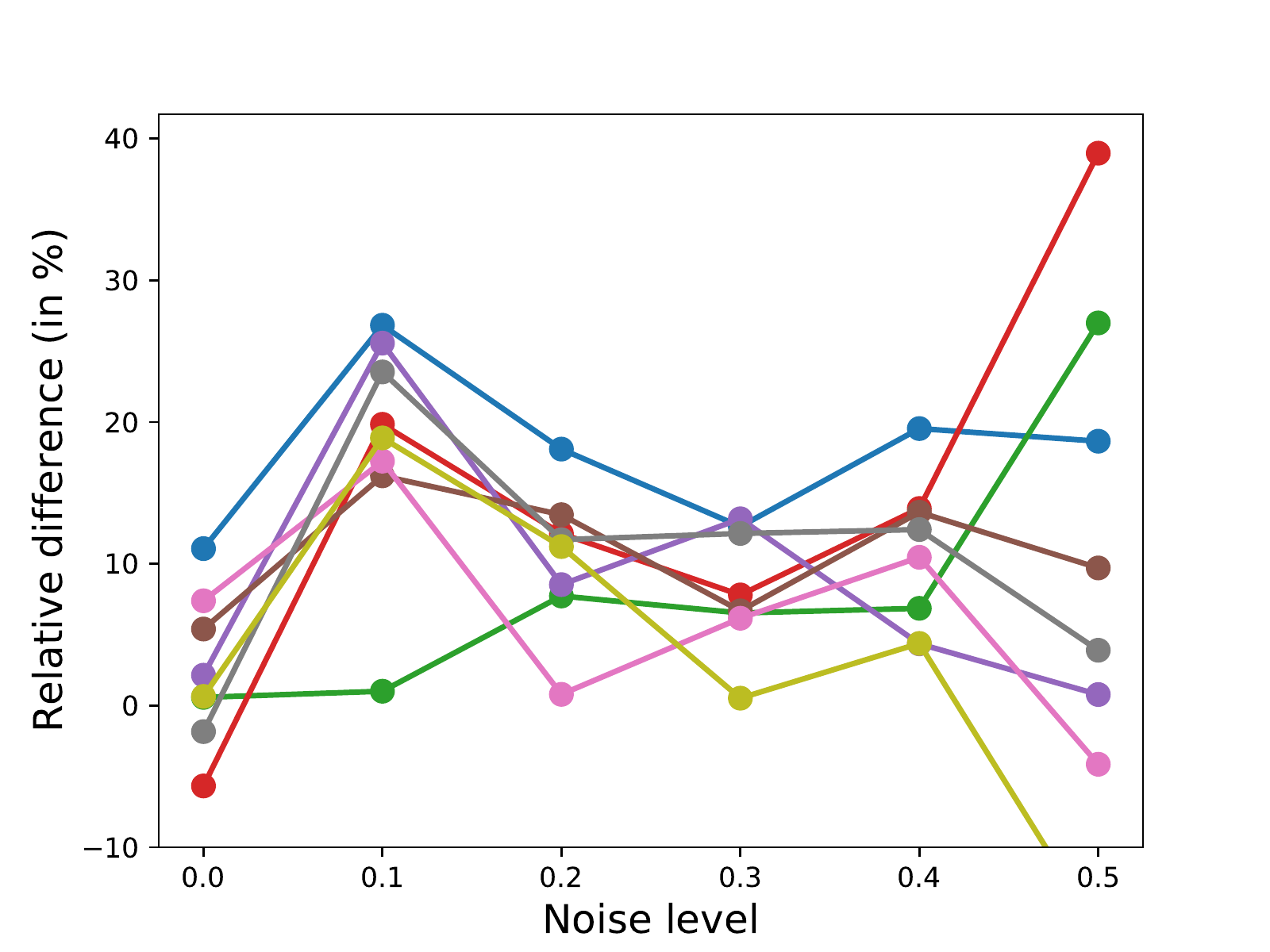}
\caption{GRSS\_DFC\_2018}
\end{subfigure}
\caption{Relative difference in performance of our proposed method and state-of-the-art methods with respect to \texttt{cross entropy} loss function for the three hyperspectral datasets: (a) Pavia University, (b) Chikusei, (c) DFC 2018 .}
\label{fig:hsi_relative}
\end{figure*}

Figure~\ref{fig:hsi_relative} shows the relative difference (in \%) of methods with respect to the baseline (\texttt{Cross entropy}) method.  The proposed \texttt{CLEOT} consistently outperformed the existing SoA methods with large margin (about 15-20\%) in the higher level label noise (except in GRSS\_DFC\_2018). However our method still has the large performance margin with other methods. On the lower level label noise, fig. \ref{fig:hsi_relative} reveals there is no significant difference between the best SoA and \texttt{CLEOT} method. However our method has several distinct advantages over the SoA, \texttt{CLEOT} (i) converges faster than the best SoA method, which is beneficial for very large scale remote sensing datasets, (ii) consistently performs better irrespective of noise level, whereas the best SoA varies with respect to the noise level, for instance with Pavia University, \texttt{Sigmoid} outperforms in mid level noise, but in higher level noise \texttt{Unhinged} outperformed the \texttt{Sigmoid} loss function, (iii) monotonically degrades the classification accuracy as the noise level increases with complex dataset, whereas SoAs do not follow this trend, thus existing methods are not as reliable.  This also reveals that the best SoAs might be more sensitive to the neural network initialization under label noise. Thus, our proposed method can be considered as a alternative candidate to train robust deep neural networks for remote sensing image analysis. 

As observed with aerial scene classification, the the classification accuracy of loss \texttt{Cross entropy} decreases  as the noise level increases. The magnitude of decrease in accuracy is dependent on the amount classes affected by the label noise, and also the nature of the datasets. It is noted that on the \emph{Chikusei} dataset, \texttt{Cross entropy} is very robust compared to other datasets, this might be due to the large patches of homogeneous landscapes in the dataset as well as the appearance of label noise at pixel level. In future work, we will consider more complicated noise model for the hyperspectral datasets, where the label noise could appears as spatially correlated clusters of pixels.

\section{Conclusion}
\label{sec7}
In this paper, we proposed the \texttt{CLEOT} method to learn robust deep neural networks under label noise in remote sensing. The proposed method leverages on the geometric structure of underlying data, and uses optimal transport with entropic regularization to regularize the classification model. We evaluated the robustness of \texttt{CLEOT} on two very different applications, one focusing on image scene classification, the second one on pixel-wise classification of hyperspectral images with  different  deep learning architectures. Our proposed approach performed better than competing state of the art approaches and has shown strong robustness in the presence of significant amount of label noise. Future works will consider other regularization schemes of the optimal transport problem, and use an embedding metric in the definition of the cost matrix $\mathbf{D}_f$ instead of relying to the distance in the input space.

\section*{Acknowledgments}
This work benefited from the support of Region Bretagne grant and  OATMIL ANR-17-CE23-0012 project of the French National Research Agency (ANR), and from CNRS PEPS 3IA DESTOPT. The authors would like to thank Prof. Paolo Gamba for providing Pavia University dataset, and the National Center for Airborne Laser Mapping and the Hyperspectral Image Analysis Laboratory at the University of Houston for acquiring and providing the GRSS\_DFC\_2018 data, and the IEEE GRSS Image Analysis and Data Fusion Technical Committee. 
We also gratefully acknowledge the support of NVIDIA Corporation with the donation of the Titan Xp GPU used for this research.

\bibliographystyle{model2-names}
\bibliography{refs}

\end{document}